\documentclass{article}
\usepackage{spconf,amsmath,graphicx}
\usepackage{amsfonts}
\usepackage{amsmath}
\usepackage{adjustbox}
\usepackage{algorithmicx}
\usepackage{algpseudocode}
\usepackage[ruled]{algorithm}
\usepackage{graphicx}
\usepackage{caption}
\usepackage{graphicx,times,amsmath} 
\usepackage{caption}
\usepackage{subcaption}
\usepackage[rightcaption]{sidecap}
\usepackage{caption}
\usepackage{multirow}
\usepackage{hhline}
\usepackage{amsmath}
\usepackage{epstopdf}
\usepackage{tabularx}

\graphicspath{ {images/} }

\usepackage{alphalph}

\title{ Generalization of Deep Neural networks for chest pathology classification in X-rays using generative adversarial networks }

\name{Hojjat~Salehinejad$^{\star \dagger}$,~Shahrokh~Valaee$^{\star}$, Tim~Dowdell$^{\dagger}$, Errol Colak$^{\dagger}$,~and~Joseph~Barfett$^{\dagger}$\thanks{The authors thank the support of NVIDIA Corporation with the
donation of the Titan X GPUs used for this project.}}
\address{$^{\star}$Department of Electrical \& Computer Engineering, University of Toronto, Toronto, Canada \\
$^{\dagger}$Department of Medical Imaging, St. Michael's Hospital, University of Toronto, Toronto, Canada \\
\textit{salehinejadh@smh.ca, valaee@ece.utoronto.ca, \{dowdellt,colake,barfettj\}@smh.ca}}

\usepackage{fancyhdr}
\fancypagestyle{pageStyleOne}{%
\normalfont\footnotesize
    \fancyhf{}
\fancyhead[L]{\fontsize{8}{8} \selectfont This paper is accepted for presentation at IEEE International Conference on Acoustics, Speech and Signal Processing (IEEE ICASSP), 2018.}
}

\begin{document}
\newcommand*{\img}{%
  \includegraphics[
    width=\linewidth,
    height=20pt,
    keepaspectratio=false,
  ]{example-image-a}%
}

\maketitle
\thispagestyle{pageStyleOne}

\begin{abstract}



Medical datasets are often highly imbalanced with over-representation of common medical problems and a paucity of data from rare conditions. We propose simulation of pathology in images to overcome the above limitations. Using chest X-rays as a model medical image, we implement a generative adversarial network (GAN) to create artificial images based upon a modest sized labeled dataset. We employ a combination of real and artificial images to train a deep convolutional neural network (DCNN) to detect pathology across five classes of chest X-rays. Furthermore, we demonstrate that augmenting the original imbalanced dataset with GAN generated images improves performance of chest pathology classification using the proposed DCNN in comparison to the same DCNN trained with the original dataset alone. This improved performance is largely attributed to balancing of the dataset using GAN generated images, where image classes that are lacking in example images are preferentially augmented.

\end{abstract}
\begin{keywords}
Chest X-ray, data augmentation, deep convolutional neural network (DCNN), generative adversarial network (GAN), simulated images.
\end{keywords}
\section{Introduction}
\label{sec:intro}
In the medical domain, preservation of patient privacy is paramount, and hence access to data is often intrinsically limited to research groups~\cite{bertino2005privacy}, \cite{salehinejad2018image}.
Medical datasets, similar to financial~\cite{salehinejad2016customer} and genomics~\cite{pouladi2015recurrent} datasets, are also very limited because they are often imbalanced~\cite{li2010learning}. Such imbalances in datasets
potentially make the training of neural networks with equally
high accuracy across classes technically challenging. Some medical problems are commonly encountered in hospital settings which leads to a substantial amount of data associated with them. However, rare conditions or syndromes such as Birt-Hogg-Dube syndrome are expected to have limited amounts of data in clinical databases~\cite{nickerson2002mutations}. 

The challenge of image availability across classes may be partially met by data augmentation techniques, for example, applying transformations to images to augment the dataset. The importance of balancing datasets is highlighted by the fact that deep neural networks may be most valuable in the work up of rare or challenging diseases, which practitioners at a common skill level may fail to recognize or misinterpret~\cite{mazurowski2008training}. Generative adversarial networks (GANs) have shown to effectively generate artificial data indiscernible from their real counterparts~\cite{tang2017automatic}. Some examples are statistical parametric speech synthesis~\cite{kaneko2017generative}, learning representations of emotional speech~\cite{7952656}, noise reduction in low-dose computed tomography (CT)~\cite{wolterink2017generative}, and retinal image synthesis~\cite{costa2017end}.

We propose the simulation of medical pathology in images as a means of augmenting data in a controlled fashion. Simulated data can be used to increase the number of images available and hence provide a means to balance datasets for the training of deep neural networks. An ideal data simulation scheme would be capable of generating an arbitrary number of synthetic images, which mimic the features of real images in any given class with sufficient diversity for the successful training of a deep network. In this manuscript, we propose the use of a deep convolutional generative adversarial network (DCGAN) for the generation of chest X-rays that mimic common chest pathologies. The synthetic images are used to balance and augment a labeled set of chest X-rays for the training of the proposed deep convolutional neural network (DCNN) across five pathological classes.

\begin{figure}[!tp]
\centering
\captionsetup{font=small}
                \includegraphics[width=0.4\textwidth]{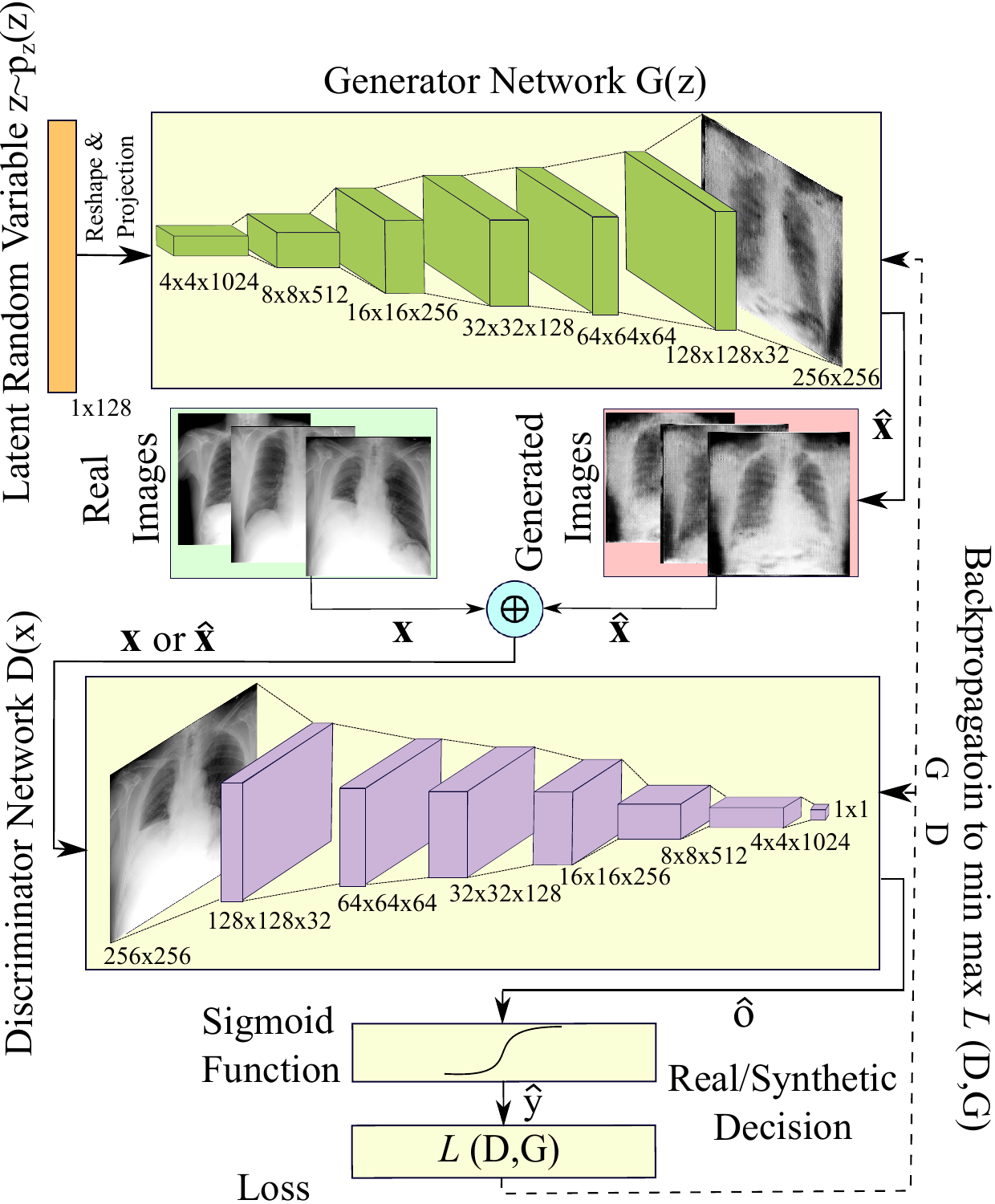}       
        \caption{Architecture of the DCGAN and training it with real chest X-rays.}
                        \label{fig:dcgan}
  \vspace{-1.5em}       
\end{figure}
\vspace{-3mm}
\section{Proposed Method}
\label{sec:proposedmethod}
\begin{figure}[!tp]
\centering
\captionsetup{font=small}
                \includegraphics[width=0.49\textwidth]{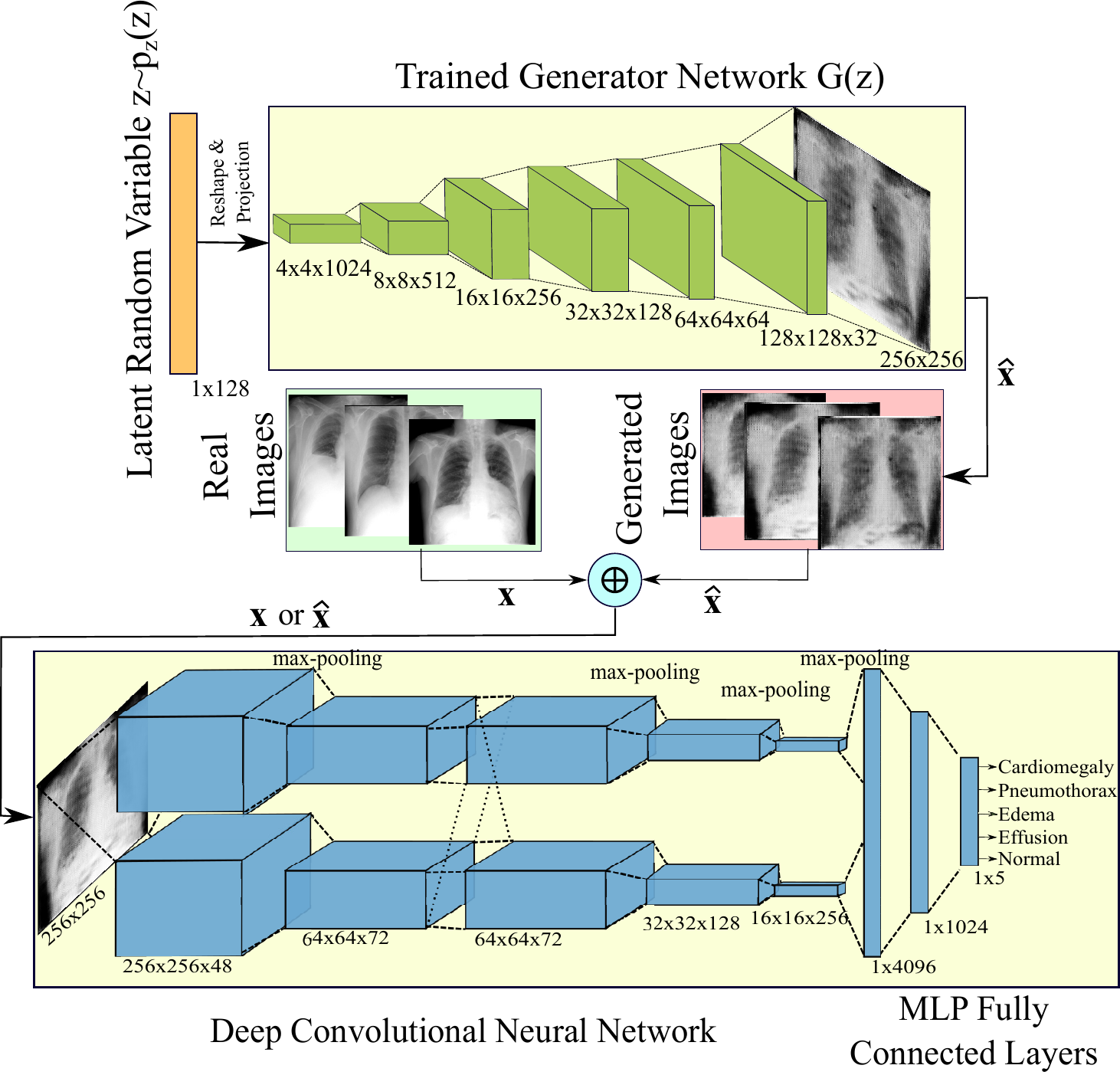}
        \caption{Architecture of the DCNN and its training with real and generated chest X-rays from DCGAN to classify abnormalities.}
\label{fig:dcnn}
  \vspace{-1.5em}       
\end{figure}

We propose a DCGAN tailored for training with chest X-rays as in Figure~\ref{fig:dcgan}. The generated artificial chest X-rays are concatenated with the real X-rays to balance and expand the training dataset to train the proposed DCNN. This DCNN performs chest pathology classification as is demonstrated in Figure~\ref{fig:dcnn}. These models are discussed in details below.

\subsection{Generating Chest X-Rays}

GANs are composed of two neural networks, a Generator $G$ and a Discriminator $D$, which compete with each other over the available training data to improve their performance. The DCGAN generates chest X-rays using DCNNs for both the $G$ and $D$ components of the model~\cite{radford2015unsupervised}. 

The trained $G$, models the underlying probability distribution $\textbf{p}_{g}$ of the training data for the set of exported chest X-rays $\textbf{x}$ and proposes artificial mappings $G(\textbf{z},\theta_{g})$ from the prior input noise variable  $p_{z}(\textbf{z})$, where $\theta_{g}$ is the set of learning parameters of the DCNN in the Generator. As demonstrated in Figure~\ref{fig:dcgan}, a 128 dimensional vector $\textbf{z}$ such that $z_{i}\sim uniform(-1,1)$ is projected to a spatially extended convolutional representation with 1,024 feature maps. Since chest X-rays contain subtle features, high resolution images are mostly of interest for machine learning purposes in medical imaging. Therefore, a series of six fractionally-strided convolutions (instead of four convolutions~\cite{radford2015unsupervised}) convert the projected and reshaped noise vector $\textbf{z}$ into a $256\times 256$ pixel chest X-ray $\hat{\textbf{x}}$.

The Discriminator network $D(\textbf{x},\theta_{d})$ receives a generated image $\hat{\textbf{x}}$ or a real chest X-ray $\textbf{x}$ and after passing that through six convolution layers, as presented in Figure~\ref{fig:dcgan}, produces an output $\hat{o}$, stating whether the input image is real or synthesized such that
\vspace{-3mm}
\begin{equation}
\hat{y}=\frac{1}{1+e^{-\hat{o}}} \:\: s.t. \:\: \hat{y}\in [0,1]
\vspace{-2mm}
\end{equation}
where $\hat{y}\to0$ and $\hat{y}\to1$ state that the input chest X-ray is synthesized or real, respectively. 
The Generator network $G$ trains so as to propose artificial images that the Discriminator network $D(\textbf{x})$ cannot distinguish from real images. The adversarial competition between $G$ and $D$ can be represented as
\vspace{-1mm}
\begin{equation}
\begin{aligned}
\min_{G} \max_{D} \mathcal{L}(D,G) = \;\:\:\:\:\:\:\:\:\:\:\:\:\:\:\:\:\:\:\:\:\:\:\;\:\:\:\:\:\:\:\:\:\:\:\:\:\:\:\:\:\:\:\:\:\:\;\:\:\:\:\:\:\:\:\:\:\:\:\:\:\:\\\ 
\mathop{\mathbb{E}}_{\textbf{x}\sim \textbf{p}_{data}}(\textbf{x})[log D(\textbf{x})]+
\mathop{\mathbb{E}}_{\textbf{z}\sim \textbf{p}_{\textbf{z}}(\textbf{z})}[log(1-D(G(\textbf{z})))]
\end{aligned}
\vspace{0mm}
\end{equation}
where the Discriminator maximizes the loss value $\mathcal{L}(D,G)$ while the Generator tries to minimize it. To train $D$, $G$ works in a feed forward fashion without back propagation, and vice versa to train $G$~\cite{goodfellow2014generative}. 

\begin{figure*}[!htp]
\centering
\captionsetup{font=small}
               
           \begin{subfigure}[t]{0.09\textwidth}
        \centering
                \includegraphics[width=1\textwidth]{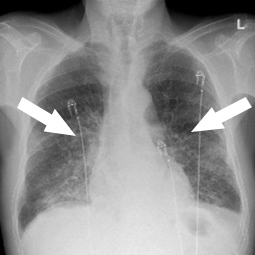}
                \caption{ }
                \label{fig:loss}
        \end{subfigure} 
        \hspace{-2mm}
               \begin{subfigure}[t]{0.09\textwidth}
        \centering
                \includegraphics[width=1\textwidth]{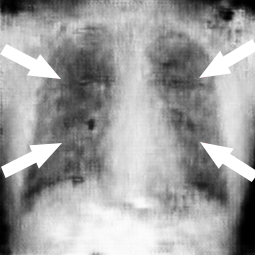}
                \caption{ }
                \label{fig:loss}
        \end{subfigure} 
        ~
           \begin{subfigure}[t]{0.09\textwidth}
        \centering
                \includegraphics[width=1\textwidth]{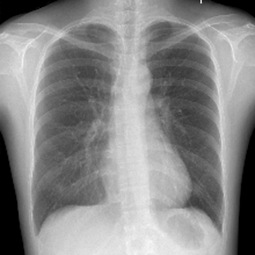}
                \caption{}
                \label{fig:loss}
        \end{subfigure} 
                \hspace{-1.8mm}
                       \begin{subfigure}[t]{0.09\textwidth}
        \centering
                \includegraphics[width=1\textwidth]{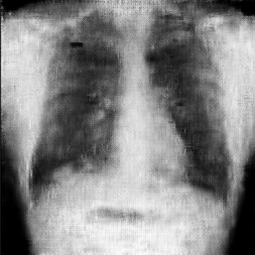}
                \caption{}
                \label{fig:loss}
        \end{subfigure}  
        ~
                         \begin{subfigure}[t]{0.09\textwidth}
        \centering
                \includegraphics[width=1.0\textwidth]{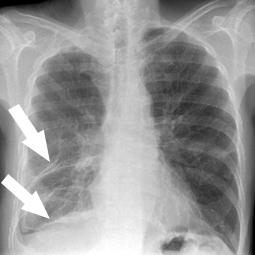}
                \caption{}
                \label{fig:loss}
        \end{subfigure}
                \hspace{-1.9mm}
                        \begin{subfigure}[t]{0.09\textwidth}
        \centering
                \includegraphics[width=0.99\textwidth]{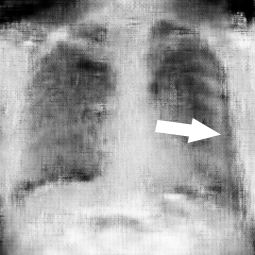}
                \caption{}
                \label{fig:loss}
        \end{subfigure}      
        ~  
        \begin{subfigure}[t]{0.09\textwidth}
        \centering
                \includegraphics[width=1\textwidth]{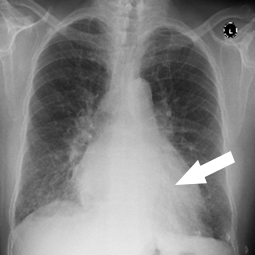}
                \caption{}
                \label{fig:loss}
        \end{subfigure}   
                \hspace{-1.5mm}
              \begin{subfigure}[t]{0.09\textwidth}
        \centering
                \includegraphics[width=1\textwidth]{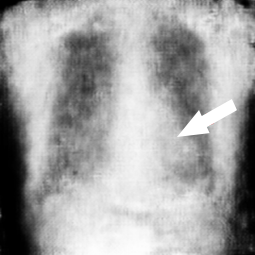}
                \caption{}
                \label{fig:loss}
        \end{subfigure}  
        ~ 
               \begin{subfigure}[t]{0.09\textwidth}
        \centering
                \includegraphics[width=1\textwidth]{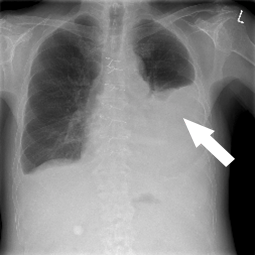}
                \caption{}
                \label{fig:loss}
        \end{subfigure}   
                \hspace{-1.5mm}
               \begin{subfigure}[t]{0.09\textwidth}
        \centering
                \includegraphics[width=1\textwidth]{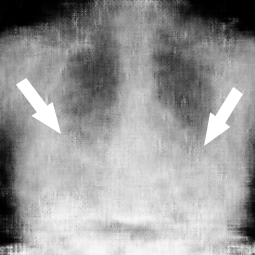}
                \caption{}
                \label{fig:loss}
        \end{subfigure}  
         \caption{Samples of real (R) and synthesized (S) chest X-rays: (a) Pulmonary Edema-R; (b) Pulmonary Edema-S; (c) Normal-R; (d) Normal-S; (e) Pneumothorax-R; (f) Pneumothorax-S; (g) Cardiomegaly-R; (h) Cardiomegaly-S; (i) Pleural Effusion-R; (j) Pleural Effusion-S. The white arrow points to the pathologic condition.}        
        \label{fig:chest_images} 
         \vspace{-4mm}
\end{figure*}

\subsection{Classification of Chest X-Rays}
The trained Generator of the discussed DCGAN is used to generate artificial chest X-rays as demonstrated in Figure~\ref{fig:dcnn}. The shuffled concatenation of real and synthesized chest X-rays is then fed into the proposed DCNN for detection and classification of pathology in chest X-rays.

AlexNet is a successful DCNN architecture that is composed of five convolutional layers for feature extraction followed by three fully-connected layers for classification~\cite{krizhevsky2012imagenet}. The proposed DCNN for chest pathology classification in this paper is fundamentally similar to AlexNet, however, uses different kernel sizes, feature map sizes, and convolution layers as illustrated in Figure~\ref{fig:dcnn}. For an input chest X-ray $\textbf{x}_{256\times 256}$, a convolution kernel of size $5\times 5$ performs the convolution operation to generate the feature map $m$ as
\begin{equation}
\vspace{-2mm}
h^{(m)}_{u,v}=\sigma(\sum_{i=0}^{4}\sum_{j=0}^{4}x_{i,j}\cdot w_{i,j}^{(m)}+b_{u,v}^{(m)}), 
\vspace{0mm}
\end{equation}
for $u \in \{0,...,255\}$ and $v \in \{0,...,255\}$, where $\sigma(\cdot)$ is the rectified linear unit (ReLU) activation function, $\textbf{b}^{(m)}$ is the bias vector, and $\textbf{w}^{(m)}$ is the weight matrix. The ReLU is defined as $f(x) = max(0, x)$, which takes advantage of its non-saturating and non-linear properties as well as the tendency to enable more efficient learning than \textit{tanh} or \textit{sigmoid} activation functions~\cite{krizhevsky2012imagenet}. A max-pooling layer after the convolution layer down-samples the latent representation by a constant factor, usually taking the maximum value over non-overlapping sub-regions such as
\vspace{-2mm}
\begin{equation}
O_{i,j}=max\{h_{q,r}^{(m)}\}, 
\vspace{0mm}
\end{equation}
for $q,r \in \{(2i,2j),(2i+1,2j),(2i,2j+1),(2i+1,2j+1)\}$, where the max-pooling kernel is a square with side length of $L=2$. This operation helps to obtain translation-invariant representations. The max-pooled features, after the multiple convolutional and max-pooling layers, are reshaped as a vector, $\textbf{f}$, and are fed to a multi-layer perceptron (MLP) network with one hidden layer as illustrated in Figure~\ref{fig:dcnn}. The output layer of MLP for the input vector $\textbf{f}$ is
\vspace{-2mm}
\begin{equation}
y_{c} = \phi(\sum_{j=0}^{|\textbf{f}|-1}f_{j}\cdot w_{j,c}+b_{c})
\vspace{-2mm}
\end{equation}   
where $|\textbf{f}|=4,096$ is the length of vector $\textbf{f}$ and $\textbf{w}$ is the weight of connections between the layers containing $\textbf{f}$ and the  output layer $\textbf{y}$. The \textit{softmax} function $\phi(\cdot)$ assigns a probability to each output unit, which corresponds to each class $c$, such as
\vspace{-2mm}
\begin{equation}
\phi(h_{c})=\frac{e^{h_{c}}}{\sum_{j=1}^{P}e^{h_{j}}}
\vspace{-2mm}
\end{equation}
where $c\in\{$Pneumothorax,  Pulmonary Edema, Pleural Effusion, Normal, Cardiomegaly$\}$, $h_{c}$ is the input to the output layer, and $C=5$ is the number of chest X-ray classes~\cite{cicero2017training}. 

\section{Experiments}
\label{sec:experiments}
We discuss the details of implementation and our obtained results in this section. 

\subsection{Data}
With the approval from our institutional ethics review board, search of our hospital's Radiology Information System (RIS) was undertaken using the Montage Search and Analytics engine. 
The dataset contained 15,781 Normal exams, 17,098 examples of Cardiomegaly, 14,510 Pleural Effusions, 5,018 examples of Pulmonary Edema, and 4,013 examples of Pneumothorax.
These images were exported and anonymized as PNG files after down-sampling to 256$\times$256 pixels to have a balance between preserving resolution and computational complexity (i.e. number of free parameters) of the models. For all the experiments, 1,000 real chest X-rays with equal contribution from each class are selected for validation and the same number for testing of the model.

\subsection{Technical Details of Training}

A team of radiologists removed inappropriate generated chest X-rays from the respective class directories and in order to keep the training dataset balanced across different classes, the DCGAN was trained with a dataset of 2,000 chest X-rays per class. The parameters of the models were as follows: the mini-batch size was 64, the number of training iterations was 20, an Adam optimizer was implemented with adaptive learning rate, starting between $2\times 10^{-4}$ and $2\times 10^{-3}$ depending on the dataset size, and momentum of 0.5. The proposed DCNN for pathology classification was trained over 100 iterations using a mini-batch size of 128, an Adam optimizer with sigmoid decay adaptive learning rate starting at  $1\times 10^{-3}$ and a momentum term of 0.5. The weights of the convolutional layers were selected using a normal distribution and biases of 0.1. The ReLU activation functions were implemented before the max-pooling layers. $L_{2}$ regularization was set to $1 \times 10^{-4}$ with early-stopping. 
Cross validation was performed 10 times prior to reporting results.

\subsection{Performance Evaluation}
Generated images were evaluated by qualitative and quantitative means. First, a board certified radiologist reviewed the images for features appropriate to the five defined classes. Second, the proportion of real and artificial chest X-rays were varied to create concatenated datasets and the results of the trained DCNN model were assessed on the test data.

\subsubsection{Qualitative Evaluation by a Radiologist}
Real and artificial images were presented to a radiologist as shown in Figure~\ref{fig:chest_images} and were visually examined. The artificial and real chest X-rays show similar characteristics, though the synthetic ones are of comparatively low resolution. For example in Figure~\ref{fig:chest_images} we demonstrate increased attenuation of the lung parenchyma in an example of pulmonary edema, a pleural line in the case of pneumothorax, an enlarged cardiac silhouette in the case of cardiomegaly, as well as bilateral pleural effusions. The generated images convincingly demonstrate the range of pathology under examination and can be classified for the training of a deep neural network. 

\begin{figure}[!t]
\centering
\captionsetup{font=small}
                \includegraphics[width=0.38\textwidth]{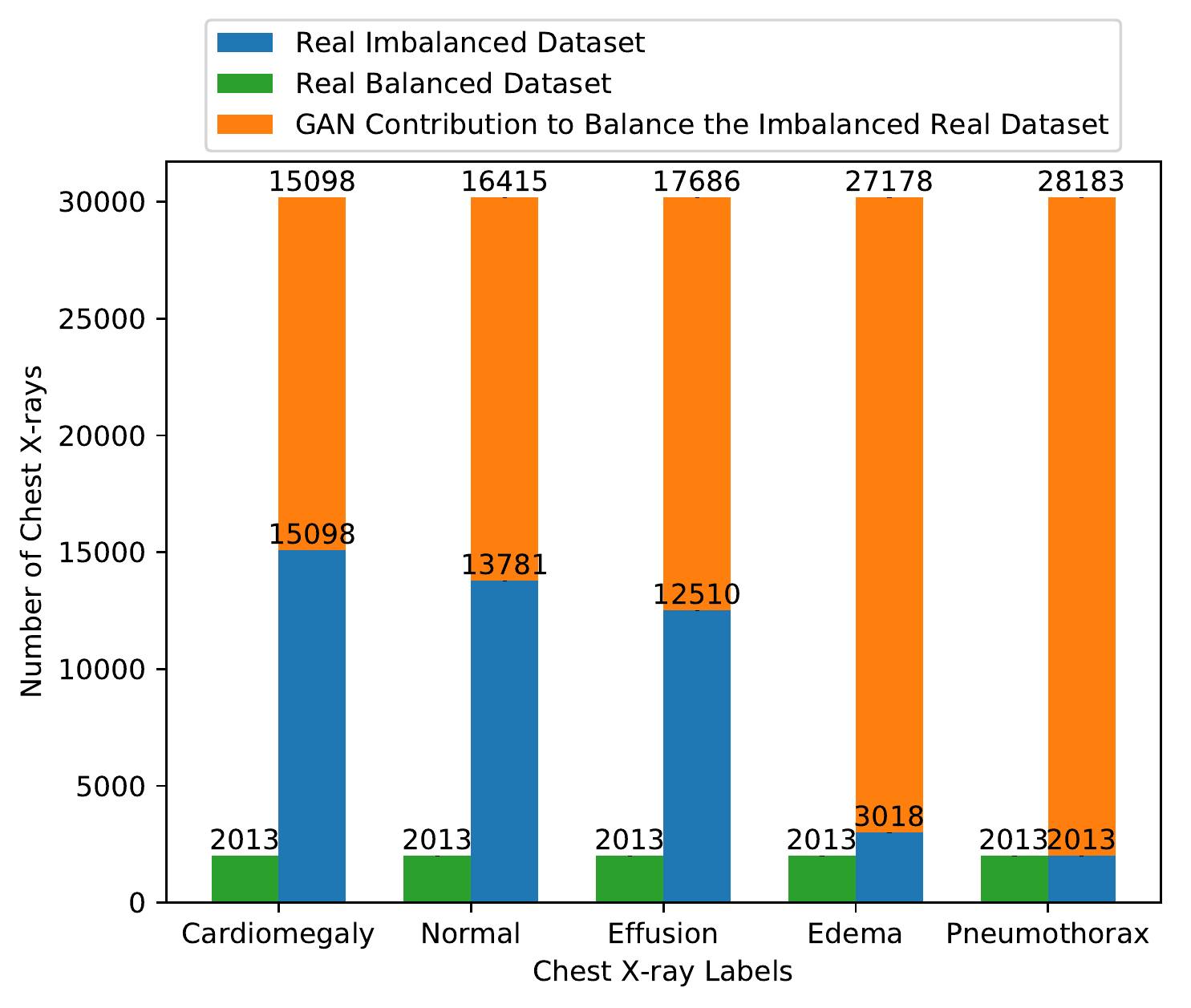}
                \caption{Number of samples for imbalanced dataset of real chest X-rays, balanced dataset of real chest X-rays, and balanced dataset of real chest X-rays using DCGAN synthesized chest X-rays. The total number of chest X-rays per label in the balanced dataset is the summation of corresponding blue and orange bars (i.e. 30,196).}
                \label{fig:stats}    
                \vspace{-0.5cm}
\end{figure}

\subsubsection{Chest Pathology with DCGAN Generated X-Rays}
Deep neural networks trained with a combination of real and artificial data have potential advantage over networks trained with real data alone, including a larger quantity of data as well as a better diversified dataset. 
Our original hospital dataset of chest X-rays, like many clinical datasets, is highly imbalanced and dominated by normal cases and common clinical conditions. Rare conditions by virtue of their low prevalence are underrepresented in such datasets.

We trained the proposed DCNN using a real imbalanced dataset (DS1), real balanced dataset (DS2), and augmented dataset with DCGAN synthesized chest X-rays to balance the imbalanced real dataset (DS3). As Figure~\ref{fig:stats} shows, each image class in the balanced dataset had 30,196 samples, which was twice the maximum number of available samples in the real imbalanced dataset (i.e. for Cardiomegaly). The accuracy of predictions for the DCNN is presented in Table~\ref{T:GANDCNN}. The augmented dataset DS3 outperforms the original dataset across all classes. A mean classification accuracy of $92.10\%$ was achieved using the proposed method, where data was augmented by DCGAN generated images to train the proposed DCNN. This performance is more than $20\%$ higher than that obtained by the same DCNN architecture trained only by the original data. The greatest performance improvement was seen in the pneumothorax class, which was the class with the fewest native images.  

\begin{table}[t]
\centering
\renewcommand{\tabcolsep}{1pt}
\captionsetup{font=small}
\footnotesize
\caption{Classification accuracy of the DCNN is improved by balancing the original dataset with DCGAN generated chest X-rays. DS1: Imbalanced dataset of real radiographs; DS2: Balanced dataset of real X-rays; DS3: Balanced dataset of real images with synthesized X-rays.}
\label{T:GANDCNN}

\begin{tabular}{|c|c|c|c|}
\hline
Accuracy (\%)    & DS1 & DS2&DS3 \\ \hline\hline
Cardiomegaly     & 79.15   & 71.73            & 95.31            \\ \hline
Normal           & 77.75    & 72.53            & 95.02            \\ \hline
Pleural Effusion & 73.64      & 51.23          & 91.19            \\ \hline
Pulmonary Edema  & 65.86     & 50.12           & 89.68            \\ \hline
Pneumothorax     & 57.99         & 48.92         & 88.84            \\ \hline
\textbf{Total }           & 70.87$\pm$0.47 & 58.90$\pm$0.48                         & \textbf{92.10$\pm$0.41}   \\ \hline
\end{tabular}
\vspace{-2mm}
\end{table}

\begin{figure}[!tp]
\centering
\captionsetup{font=small}
                \includegraphics[width=0.4\textwidth]{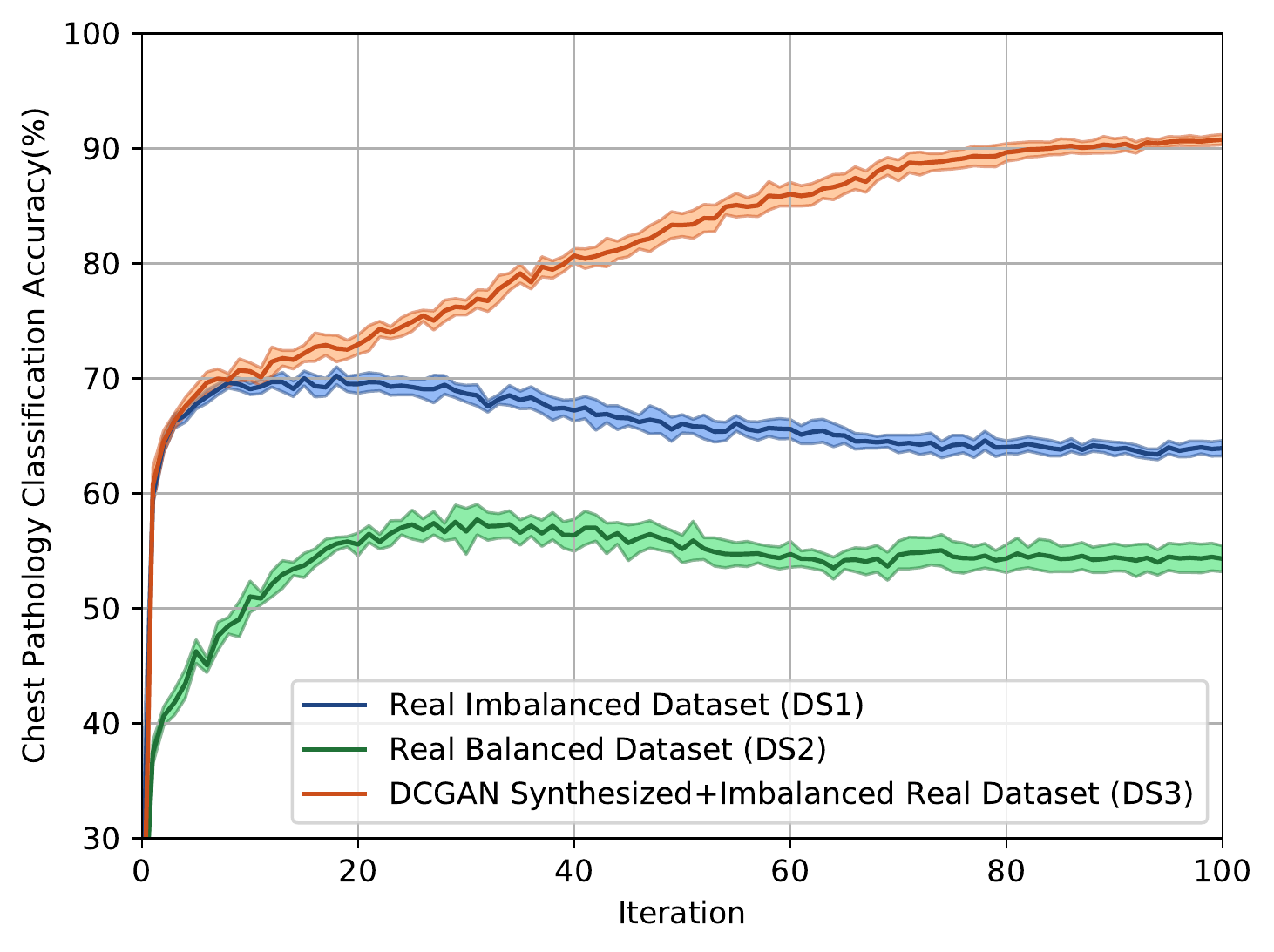}
               
        \caption{Accuracy (solid line) and standard deviation (shaded region) of the DCNN on the validation dataset during training iterations over the DS1, DS2, and DS3 datasets.}
\label{fig:validation_plots}
                \vspace{-0.5cm}
\end{figure}

Accuracy and standard deviation of the proposed DCNN on the validation dataset for training
iterations over DS1, DS2, and DS3 datasets are presented in Figure~\ref{fig:validation_plots}. The DCNN trained with DS3 has achieved higher accuracy comparing with the DS1 and DS2.
The plot show that the DCNN trained on DS3 has almost
converged after 90 iterations. The same DCNN trained on 
DS1 and DS2 is overfitted after 22 and
31 iterations, respectively, to a lower accuracy. The plot clearly shows
effectiveness of the added diversity to the dataset by the generated
chest X-rays using the DCGAN, which helps to improve generalization performance of the DCNN for chest pathology classification in X-rays and avoid over-fitting.

\section{Conclusion}
\label{sec:conclusion}

In this paper, we have shown that artificial data generated by DCGAN can augment real datasets to provide both a greater quantity of data for training of large neural networks and can balance the dataset, resulting in substantial improvement in classification performance in the most anemic classes. We obtained best results with a combination of real and artificial data used to train the DCNN. The reported results in this paper suggest that data augmentation using synthesized images increases the diversity of training dataset and therefore, improves generalization performance of deep learning for classification of unseen data. 

\bibliographystyle{IEEEbib}
\bibliography{strings,mybibfile}

\end{document}